\documentclass[10pt]{cai}
\usepackage{hyperref}

\graphicspath{{figs/}}

\newcommand{\guillemet}[1]{``#1''}

\begin{document}

\volumeheader{35}{0}
\begin{center}

  \title{Quantifying French Document Complexity}
  \maketitle

  \thispagestyle{empty}

  \begin{tabular}{cc}
    Vincent Primpied\upstairs{\affilone,*}, David Beauchemin\upstairs{\affilone}, Richard Khoury\upstairs{\affilone}
   \\
   {\small \upstairs{\affilone} Department of Computer Science and Software Engineering, Université Laval} \\
  \end{tabular}
  
  \emails{
    \upstairs{*}vincent.primpied.1@ulaval.ca
    }
  \vspace*{0.2in}
\end{center}

\begin{abstract}
Measuring a document's complexity level is an open challenge, particularly when one is working on a diverse corpus of documents rather than comparing several documents on a similar topic or working on a language other than English. In this paper, we define a methodology to measure the complexity of French documents, using a new general and diversified corpus of texts, the \href{https://github.com/GRAAL-Research/FCCLC}{\guillemet{French Canadian complexity level corpus}}, and a wide range of metrics. We compare different learning algorithms to this task and contrast their performances and their observations on which characteristics of the texts are more significant to their complexity. Our results show that our methodology gives a general-purpose measurement of text complexity in French. 

\end{abstract}

\copyrightnotice

\section{Introduction}
\label{sec:intro}
The evaluation of text complexity is intuitive to human readers. Upon looking at a new document, a reader can almost immediately tell if the text is easy or difficult to read. 
However, this task is more difficult for reading comprehension systems \cite{tolochko2019determining}. 
While text complexity evaluation has evolved greatly recently in particular thanks to the rise of machine learning (ML) \cite{franccois2012nlp}, most applications focus on measuring the complexity of a specific type of text, such as political discourse \cite{tolochko2019determining} or administrative documents \cite{franccois2014amesure}, and mostly in English \cite{franccois2012ai}. As a result, most work on text complexity focuses on a specific characteristic of the language, such as the use of commas or the variety of words, which correlates with complexity in a specific context but does not generalize well for all contexts nor all languages \cite{franccois2012ai}. 

This paper explores the design and training of a general-purpose text complexity measuring algorithm in French. To achieve this goal, we pick out a variety of metrics that the literature identified as significant to complexity and train multiple learning algorithms to combine them to represent the general, non-application-specific complexity of a wide range of documents. We focus specifically on French, using our own French document corpus and metrics specialized to the French language. Our methodology can be applied to other languages as well by using a different corpus and metrics.

The rest of this paper is structured as follows. We will begin with a brief overview of the literature on document complexity in \autoref{sec:relatedwork}, and list the metrics commonly used for that task in \autoref{sec:metrics}. Then we will present our work on training a complexity evaluation algorithm in \autoref{sec:experiments}, including notably how we constructed our French corpus, the various ML algorithms we trained, the results they obtained and the features they each picked out as significant to complexity. Finally, we will make some concluding remarks in \autoref{sec:conclusion}.

\section{Related Work}
\label{sec:relatedwork}
One text-complexity measuring system is the Coh-Metrix Common Core Text Ease and Readability Assessor \cite{jackson2016common}, which computes metrics classified in 5 specific aspects of readability in literacy instruction: narrativity, syntactic simplicity, word concreteness, referential cohesion and deep cohesion. Although domain-specific, many metrics overlap with more generic work.
The authors of \cite{kyle2018measuring} add that when evaluating syntactic complexity, subdividing into clausal and phrasal indices can be very relevant in the evaluation for English learner, and in a more generic evaluation. 
The authors of \cite{tolochko2019determining} proposed an analysis for a global complexity measure of political discourse. They conceptualize text complexity as two dimensions: syntactic complexity and semantic complexity. After testing, they conclude that the complexity rating has to be based on a multi-dimensional analysis to be relevant. It is interesting to note that the same happened in the German language.

These studies were all developed for the English language
and are not compatible with a French application. The two languages have major differences, such as French being much richer in morphology \cite{de2019development}. As a result, the tools and even basic concepts used to quantify English complexity cannot be applied to French documents.

One example of a French text complexity system is AMESURE. It aims to evaluate the complexity of administrative documents and to give advice on how to reduce it \cite{franccois2014amesure}. It allows users to see exactly what caused a grade by highlighting too long or too complex sentences.
However, while it performs well in its intended domain, the system has difficulty assessing complexity in a context besides administrative texts. This is because a reader experiences text complexity according to their expectations of the document. Therefore, measuring it depends on the application domain \cite{tolochko2019determining}, and a domain-specific complexity measure makes domain-specific assumptions and design choices that limit its transferability. Moreover, AMESURE is based on a support vector machine that uses ten metrics representing three aspects of complexity, limiting its score.

\section{Metrics}
\label{sec:metrics}
One of the most intuitive concepts of text complexity is the variety of vocabulary used, or lexical diversity: the more diverse the vocabulary present in a document is, the more complex it is. Hence, one of the first metrics of text complexity created is the Type-Token Ratio (TTR) \cite{johnson1944studies, templin1957certain}, which simply measures the ratio of unique words (types) to total words (tokens) found in a document. Unfortunately, this metric is very sensitive to the length of the text: longer texts have more tokens and, therefore, a lower TTR score regardless of their actual difficulty level. 
Thus, subsequent authors have expressed the TTR differently to combat this issue \cite{covington2007mattr, mccarthy2010mtld}.
The Mean Sequential TTR (MSTTR) \cite{johnson1944studies, templin1957certain} computes the TTR on each 50-word window of the document and the result is the averages of these TTRs. The Moving Average TTR \cite{covington2007mattr} does the same on a 100-word rolling window instead of sequential windows. 
Finally, the Measure of Textual Diversity (MTLD) \cite{mccarthy2010mtld} takes the opposite approach: it sets a target TTR value (the authors recommend 0.720) and gradually increases the size of the text window until the TTR of that window falls below that threshold, at which point a counter is incremented and the window is reset. The document's MTLD value is the ratio of the number of words to the counter value.

\begin{table}
    \centering
    \begin{tabular}{lll}
    \toprule
       Name  &  Description  & Reference  \\
        \midrule
        TTR & Type-Token Ratio & \cite{johnson1944studies, templin1957certain} \\
        MSTTR & Mean Sequential TTR (50-word sequences)& \cite{johnson1944studies, templin1957certain} \\
        MATTR & Moving Average TTR (100-word window)& \cite{covington2007mattr} \\
        MTLD & Mean document length per TTR value & \cite{mccarthy2010mtld} \\
    \bottomrule
    \end{tabular}
    \caption{Summary of the lexical diversity metrics}
    \label{tab:ld_metrics}
\end{table}

Some authors note that there is more to vocabulary complexity than diversity: a document with a diverse but simple vocabulary will seem less complex than one with a limited vocabulary of difficult or technical words. The simplest way to measure this notion of complexity is to count the percentage of words not found in a list of simple words \cite{dale1948formula}. The authors of \cite{si2001statistical} take a very different approach and define a list of unigrams of words corresponding to different complexity levels, then count occurrences of these words in texts. While these were done in English, equivalent word lists are available in French, such as the Gougenheim fundamental dictionary \cite{gougenheim1958dictionnaire} or the Lexique 3 list of lemmas and their probability of occurrence \cite{new2005manuel}. Finally, going by the assumption that longer words are more difficult, several authors propose simply counting the average length of words \cite{flesch1948readability, smith1961devereux}. 
\begin{table}
    \centering
    \begin{tabular}{lp{8cm}l}
    \toprule
       Name  &  Description  & Reference  \\
       \midrule
        PA & Percentage of words not in the Gougenheim dictionary & \cite{gougenheim1958dictionnaire, dale1948formula} \\
        Unigram & Unigram of lemma sorted by Lexique3 frequency & \cite{new2005manuel, si2001statistical}\\
        NLM & Mean length of words & \cite{flesch1948readability, smith1961devereux} \\
        \bottomrule
    \end{tabular}
    \caption{Vocabulary complexity metrics}
    \label{tab:vc_metrics}
\end{table}

Another aspect of text complexity is syntactic complexity, or the use of simpler or more complex grammatical constructions. The simplest metric to measure this complexity is the Mean Length of Sentences (MLS), and its variations in the \guillemet{length of the 90th percentile sentence} or the \guillemet{percentage of sentences longer than 30 words} \cite{flesch1948readability}. 
More accurate metrics require first splitting each sentence into clauses, determining if it is dependent or not and computing the Mean Length of Clauses \cite{flesch1948readability} and the Dependent Clauses per Clause ($\frac{\text{DC}}{\text{C}}$).
An even more accurate metric has been found to be the T-unit, which is defined as the root clause of a sentence along with all its subordinate clauses except for conjunctions \cite{hunt1965grammatical}. A complex T-Unit (CTU) is a type of T-Unit with more than one clause\footnote{For example, the conjunctive sentence ``\textit{Il fait beau et les nuages sont partis.}'' (``The weather is nice and the clouds are gone.'') contains two T-units and two clauses, while the explicative sentence ``\textit{Il fait beau parce que les nuages sont partis.}'' (``The weather is nice because the clouds are gone.'') contains one complex T-unit but two clauses.}. In contrast to simply measuring clauses, using T-Units makes it possible to isolate complex grammatical constructs.

The metrics we can draw from this shortest grammatically relevant sentences and other grammatical units are listed in \autoref{tab:sc_metrics}.

\begin{table}
      \centering
        \begin{tabular}{lp{3.5cm}llp{3.5cm}l}
            \toprule
       Name  &  Description & Reference  & Name  &  Description & Reference\\
       \midrule
       MLS & Mean Length of Sentences & \cite{hunt1965grammatical} &$\frac{\text{C}}{\text{TU}}$ & Clauses per T-Units & \cite{bardovi1989attainment}\\
       MLC & Mean Length of Clauses & \cite{hunt1965grammatical} & $\frac{\text{CP}}{\text{C}}$ & Coordinate phrase per Clauses & \cite{lu2010automatic}\\
       $\frac{\text{DC}}{\text{C}}$ & Dependent Clauses per Clauses & \cite{hunt1965grammatical} &$\frac{\text{CP}}{\text{TU}}$ & Coordinate phrase per T-Units & \cite{lu2010automatic}\\
       MLT & Mean Length of T-units & \cite{hunt1965grammatical} & NWS$_{90}$ & Length of the $90^{th}$ percentile sentence & \cite{hunt1965grammatical} \\
        $\frac{\text{TU}}{\text{S}}$ & T-Unit per Sentences & \cite{hunt1965grammatical} & PS$_{30}$ & Percentage of sentences longer than 30 words & \cite{hunt1965grammatical} \\
        $\frac{\text{CTU}}{\text{TU}}$ & Complex T-Unit per T-Units & \cite{casanave1994language, lu2010automatic} & & &\\
        \bottomrule
        \end{tabular}
    \caption{Syntactic complexity metrics}
    \label{tab:sc_metrics}
\end{table}

Finally, some authors propose to quantify and measure a document's readability. The traditional way of doing this is with the popular Flesch-Kincaid reading ease formula \cite{flesch1948readability}, and his adaptation in French the Kandel and Moles formula \cite{kandel1958application}. This readability measurement is given by \autoref{form:km_formula} and combines the MLS and the Mean Number of Syllables (MNS). Other authors noted that readability is inversely related to the use of commas. The intuition is that commas are likely used to introduce additional dependent clauses in a sentence and thus make the sentence more complex and more difficult to read. The BINGUI value \cite{franccois2011apports} measures this effect. For our work, we extend this metric to include semi-colons, colons and parentheses, and commas, which all have a similar effect on readability. These readability metrics are summarized in \autoref{tab:read_metrics}.
\vspace{-0.5em}

\begin{align}
    \label{form:km_formula}
    \text{KM}_{\text{score}} = 207-1.015\times \text{MLS} - 0.736\times \text{MNS}
\end{align}

\begin{table}
    \centering
    \begin{tabular}{lp{11cm}l}
    \toprule
       Name  &  Description & Reference  \\
       \midrule
        FK ease & Flesch-Kincaid reading ease & \cite{flesch1948readability} \\
        KM score & Flesch reading ease, adapted for French by Kandels and Moles & \cite{kandel1958application} \\
        BINGUI & Commas per sentence, which we extend to include semi-colon, colon and parenthesis & \cite{franccois2011apports} \\
        \bottomrule
    \end{tabular}
    \caption{Readability metrics}
    \label{tab:read_metrics}
\end{table}

In addition to these complexity metrics, we apply the Biber tagger \cite{biber1988adverbial} to each document in order to measure a set of linguistic features present in the text. We use the implementation available through the \verb|biberpy| library \cite{sharoff21rs}, which measures the features listed in \autoref{tab:biber_tagger}. We normalize these values as the ratio of each count to the number of words in the text.

\begin{table}
    \centering
    \begin{tabular}{llll}
    \hline
        \textbf{Metric} & \textbf{Description} & \textbf{Metric} & \textbf{Description} \\
        \midrule
         pastVerbs & Past tense verb & otherSubord & Other clauses\\
         
         presVerbs & Present tense verb & preposn & Adposition\\
         placeAdverbials & Place adverbials & attrAdj & Attributive adjectives\\
         timeAdverbials & Time adverbials & ADV & Adverb\\
         1persProns & 1st person pronouns & conjuncts & Conjunct\\
         2persProns & 2nd person pronouns & downtoners & Downtoners\\
         3persProns & 3rd person pronouns & amplifiers & Amplifiers\\
         impersProns & Impersonal pronouns & generalEmphatics & Emphatics\\
         demonstrProns & Demonstrative pronouns & publicVerbs & Speech act (public) verbs\\
         indefProns & Indefinite pronouns & privateVerbs & Intellectual state (private) verb\\
         doAsProVerb & Do as pro-verb & suasiveVerbs & Suasive Verbs\\
         whQuestions & WH- questions words & seemappear & Occurences of \guillemet{seem} or \guillemet{appear}\\
         nominalizations & Nominalizations & possibModals & Possibility modals\\
         Nouns & Nouns (without nomalizations) & necessModals & Necessity modals\\
         beAsMain & Be as sentence's root & predicModals & Prediction modals\\
         WHclauses & WH relative clauses & contractions & Word contractions\\
         piedPiping & Pied-piping constructions & thatDeletion & Omission of \guillemet{that}\\
         sncRelatives & Relative clauses & strandedPrep & Preposition strandingbonjour
         \\
         causative & Causative adverbial clauses & syntNegn & Syntactic negation\\
         conditional & Conditional clauses & analNegn & Analytic negation\\\bottomrule
    \end{tabular}
    \caption{List of metrics taken from the Biber tagger.}
    \label{tab:biber_tagger}
    \vspace{-2em}
\end{table}

\section{Experiments}
\label{sec:experiments}

\subsection{Corpus}
\label{sec:corpus}
Our first challenge is that no corpus exists to study general text complexity in French. Consequently, we decided to build our own. 
We identified eight main groups of documents representing distinct points on the spectrum of writing styles and difficulty levels:
children's stories, recipes, news articles, Wikipedia articles, novels, dictations, insurance documents and legal texts. Children's stories and recipes are the two simplest forms of text we sampled: the former because it deliberately uses simpler syntax and vocabulary. The latter is because it is a set of clear and concise directions. 
Novels, news articles and Wikipedia articles are of medium complexity, as they are intended for a more knowledgeable public. Novels are a more complex form of our previous stories category. At the same time, news articles and Wikipedia articles are both texts intended to convey a lot of information to the reader, with news articles geared more toward the general public and Wikipedia articles towards individuals interested in an in-depth technical explanation. 
Dictations are intermediate between these; the text is more complex than stories and recipes but less so than novels and articles. It will also contain some outlier features intended to test the limits of the students' knowledge. 
Finally, insurance and legal documents, two forms of text that use a very technical vocabulary and complex syntactic structures, are the most difficult forms of text we sampled.
The three main tiers of complexity (children's stories and recipes/news, Wikipedia articles, novels and dictation/insurance and legal documents) are clear; however, the ranking within each tier is based on our intuition. This, however, will not affect the quality of our results in a significant manner. Future work may create a more objective tier ranking using focus groups or a set of independent labellers.
Note that we selected a maximum of texts in Canadian French instead of other French dialects whenever possible.
Our small assembled corpus, the \guillemet{French Canadian complexity level corpus}, is composed of 118 documents and is available on our \href{https://github.com/GRAAL-Research/FCCLC}{GitHub repository}.

\vspace{-1em}
\subsection{Results per Metric}
\label{sec:resultsmetric}
The first step of our study is to explore how each metric correlates with document complexity. To do this, we compute each metric for each document independently and group metrics that behave similarly. This allows us to discern three different groups of behaviours for the metrics. 

The first group is for metrics that are highly correlated with complexity but overlap in values between similar categories. These metrics are useful to get a coarse evaluation of complexity but cannot differentiate between fine-grained levels of complexity. Moreover, a metric in this group can occasionally misevaluate one level of complexity but be correlated for the other seven. For example, the MLS is only based on the length of sentences and will classify a straightforward amendment with short sentences as easy as a children's story.
 
\begin{figure}
    \centering
    \begin{minipage}{0.48\textwidth}
        \captionsetup{width=.9\linewidth}
        \centering
        \includegraphics[scale=0.45]{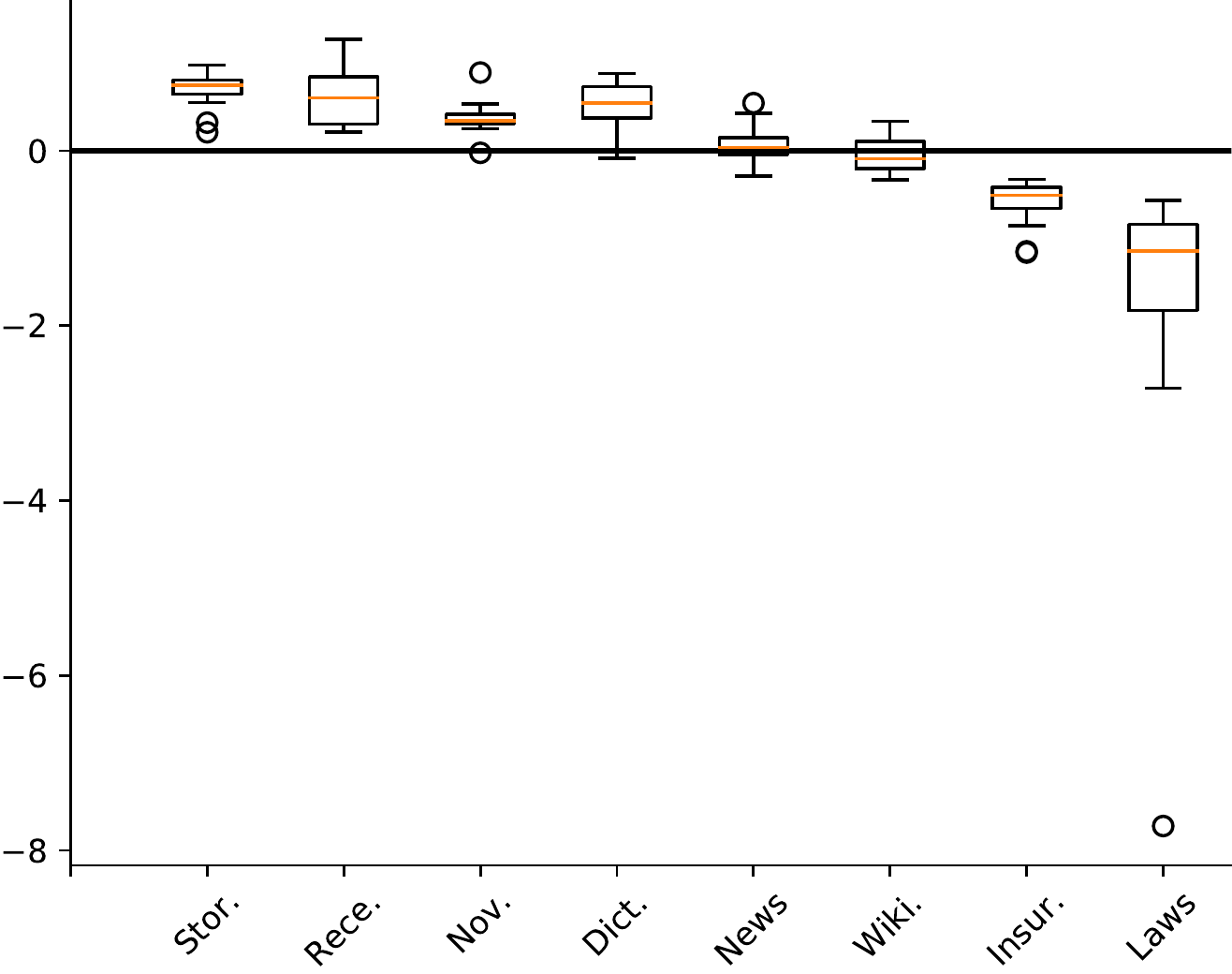}
        \caption{Box plot of the Kandel and Moles formula for each complexity level}
        \label{fig:boxplot_km}
    \end{minipage}
    \begin{minipage}{0.48\textwidth}
        \captionsetup{width=.9\linewidth}
        \centering
        \includegraphics[scale=0.45]{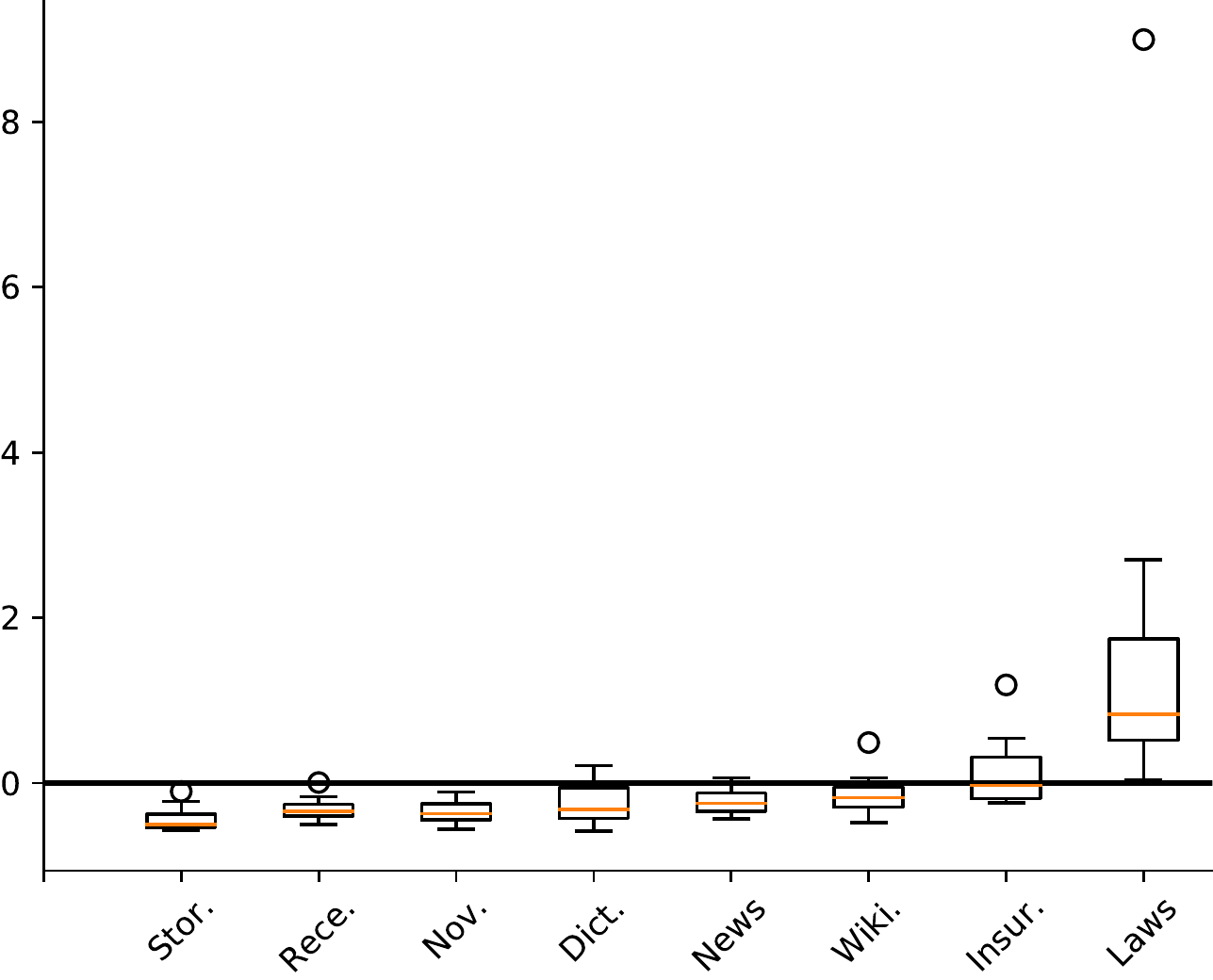}
        \caption{Box plot of the BINGUI metric for each complexity level}
       \label{fig:boxplot_bingui}
    \end{minipage}
    
    \begin{minipage}{0.48\textwidth}
        \captionsetup{width=.9\linewidth}
        \centering
        \includegraphics[scale=0.45]{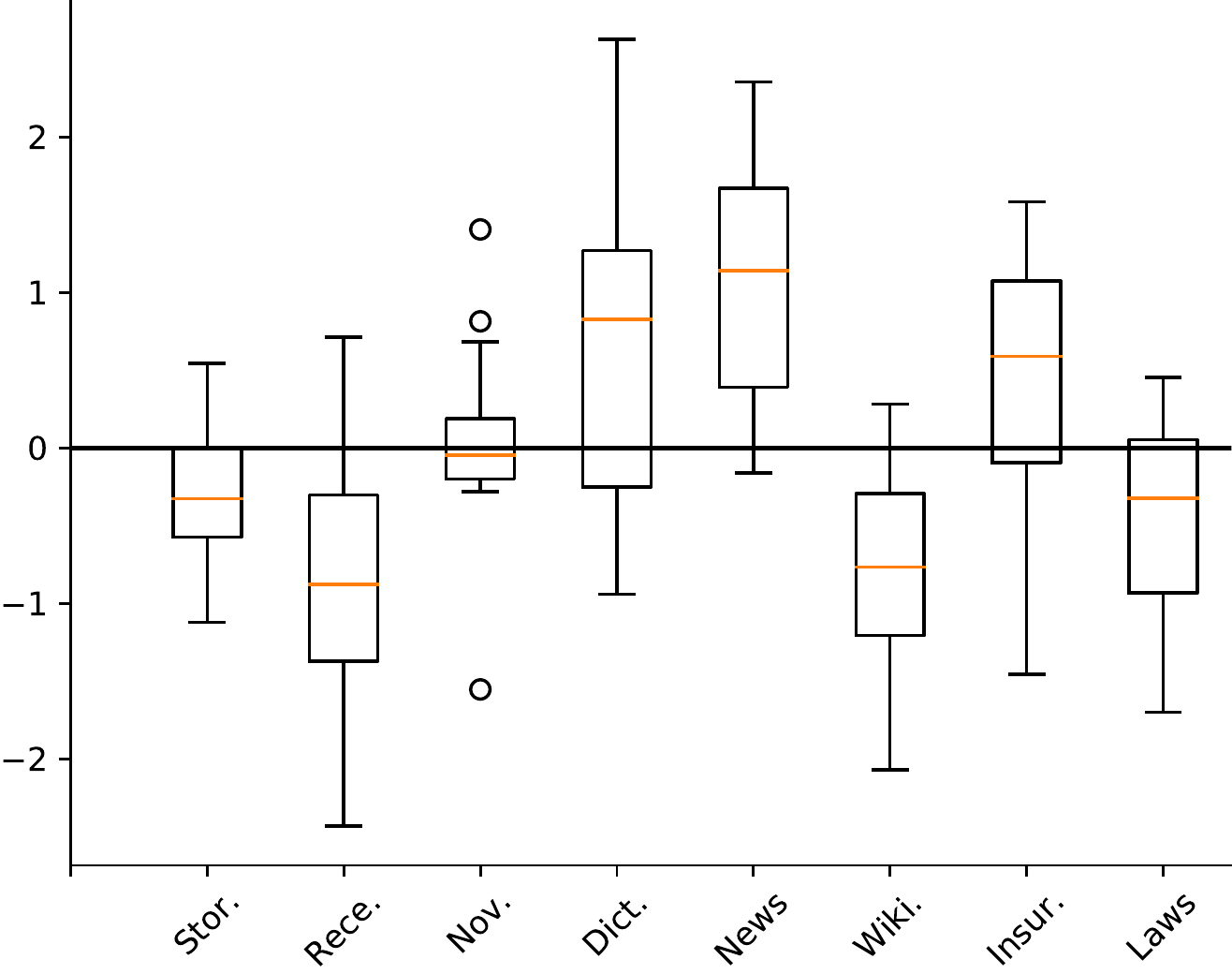}
        \caption{Box plot of the $\frac{\text{CTU}}{\text{TU}}$ metric for each complexity level}
        \label{fig:boxplot_analnegn}
    \end{minipage}
    \begin{minipage}{0.48\textwidth}
        \captionsetup{width=.9\linewidth}
        \centering
        \includegraphics[scale=0.45]{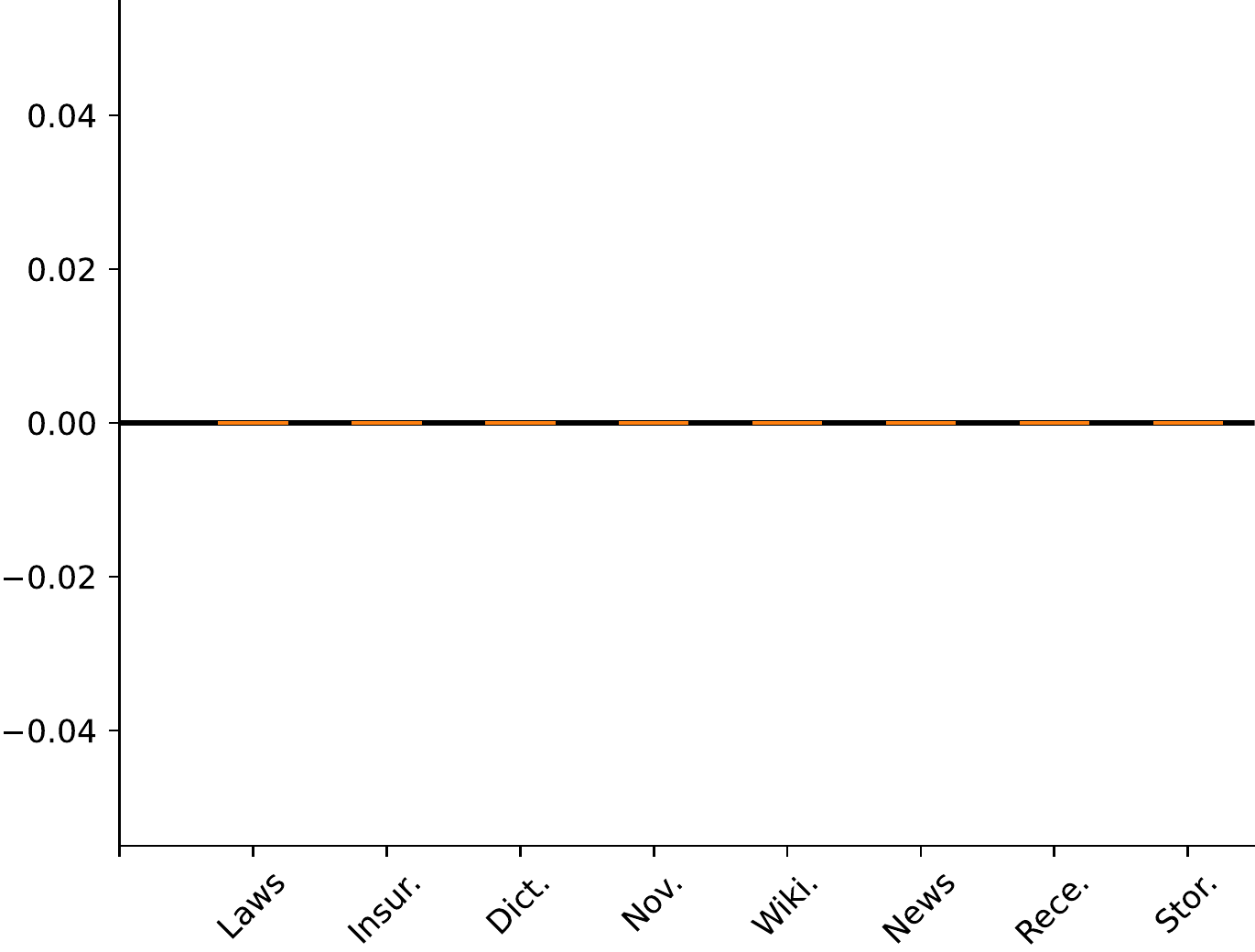}
        \caption{Box plot of a metric not applicable to French, the \textit{discoursePart}}
        \label{fig:box_plot_discourse}
    \end{minipage}
   \vspace{-1em}
\end{figure}

One typical example of metrics in this group is the Kandel and Moles formula, which we present in \autoref{fig:boxplot_km}: there is a clear progression of values from the simplest to most complex level of texts, but too much overlap between levels to reliably classify a document in one category based on that information alone. The other metrics in that group are Flesch-Kincaid reading ease, nominalizations, MLS and $\frac{\text{DC}}{\text{C}}$ .

The second group are metrics that overlap in value for most levels of complexity but give very different results for one level (or few levels).

The BINGUI metric shown in \autoref{fig:boxplot_bingui} is a good example of such a metric: it gives very different results for legal documents but reliably cannot tell the other seven levels apart. The other metrics in that group are the MSTTR, TTR, conjuncts, generalEmphatics, strandedPrep, synthNegn, possibModals, MTLD, MATTR, contractions, $\frac{\text{CP}}{\text{TU}}$, $\frac{\text{TU}}{\text{S}}$, wordLength, $\frac{\text{C}}{\text{TU}}$, pastVerbs, $PS_{30}$ and MLT.
Their usefulness is their ability to isolate one level of complexity from the others.

The final group is for metrics that do not fit in the previous two: they present a low correlation with text complexity and do not have a single level that stands out from others. 

The $\frac{\text{CTU}}{\text{TU}}$ metric shown in \autoref{fig:boxplot_analnegn} is a good representative of metrics in this group, and the others are listed in \autoref{fig:low_correlation}.
While such a metric does not seem helpful on its own, combined with other metrics, it may prove helpful. 

\begin{table}
    \centering
    \begin{tabular}{lllllll}
    \toprule
    Unigram      & ADV           & amplifiers                   & preposn      & downtoners & presVerbs & Nouns\\
    PA           & piedPiping    & NLM                          & conditionnal & demonstrProns & doAsProVerb\\
    WHclauses    & 2persProns    & timeAdverbials               & sncRelatives & analNegn & causative\\
    predicModals & impersProns   & $\frac{\text{CP}}{\text{C}}$ & indefProns   & otherSubord & $NWS_{90}$\\
    necessModals & suasiceVerbs  & whQuestions                  & 1persProns   & thatDeletion & placeAdverbial\\
    attrAdj      & 3persProns    & publicVerbs                  & seemappear   & beAsMain & privateVerbs\\
    \bottomrule
    \end{tabular}
    \caption{List of metric with a low correlation with text complexity}
    \label{fig:low_correlation}
\end{table}

The Biber tagger identifies complexity forms that have not been considered. Unfortunately, some of the concepts measured by the tagger do not apply to French. Theses are easily identifiable: \autoref{fig:box_plot_discourse} shows an example. 

\subsection{Results with Machine Learning}
\label{sec:resultslearning}
Since no metric can precisely differentiate between all eight complexity levels, the next stage of our research is to use ML to find a way to combine them and gain that ability. 
We will use a readability metric as a baseline to make a relevant comparison.

Furthermore, to train our ML models, each complexity level is labelled as a number class from 0 to 7. Zero is the easiest class (Children's stories), and seven is the most difficult (law documents).

The classifier models we experimented with are a Decision Tree (DT), Random Forest (RF), a Logistic Regression (LR), and a Naive Bayes (NB), from \verb|scikit-learn|. First, we perform a grid search with cross-validation to optimize each of our ML algorithms. For the NB, we perform the grid search on the Laplace smoothing value. For the DT, we alternate between Gini or entropy as a quality metric and optimize the maximum depth. As for the RF, we use the same parameters as the DT plus the number of estimators in the forest. Finally, for the LR, we search over the regularization strength. As a baseline, we use an implementation of the Kandel and Moles formula with thresholds \cite{franccois2012nlp, benoit1986revue}. These thresholds are associated with the definition of the Kandel and Moles formula \cite{kandel1958application}, as the \autoref{tab:km_thresholds} show. Complete code is available on our \href{https://github.com/GRAAL-Research/TextComplexityComputer}{GitHub repository}.

\begin{table}
    \centering
    \begin{minipage}{0.49\textwidth}
        \centering
        \captionsetup{width=.9\linewidth}
        \begin{tabular}{cc}
        \toprule
            Score & Description of style \\
             \midrule
            90-100 & Very easy \\
            80-90 & Easy \\
            70-80 & Fairly easy \\
            60-70 & Standard \\
            50-60 & Fairly difficult \\
            30-50 & Difficult \\
            0-30 & Very difficult \\
          \bottomrule
        \end{tabular}
        \caption{Kandel and Moles threshold according to the authors \cite{kandel1958application}}
        \label{tab:km_thresholds}
    \end{minipage}
    \begin{minipage}{0.49\textwidth}
    \centering
        \captionsetup{width=.9\linewidth}
        \centering
        \begin{tabular}{lcc}
        \toprule
            Model & RMSE & Accuracy \\
            \midrule
            Baseline & 1.28 & 0.38\\
            Decision Tree (DT) & 1.43 & 0.65\\
            Logistic Regression (LR) & 0.38 & 0.83 \\
            Multinomial Naive Bayes (NB) & 0.36 & 0.84\\
            Random Forest (RF) & \textbf{0.21} & \textbf{0.92}\\
            \bottomrule
        \end{tabular}
        \caption{RMSE and Accuracy of the baseline and our four best ML configurations (bold value are the best performance per metric)}
        \label{tab:modelRA}
    \end{minipage}
\end{table} 

We measure two metrics of the quality of the results. The first is accuracy, or the proportion of documents classified at the correct complexity level. The second is the Root-Mean-Square Error (RMSE), which is computed by assigning to each complexity level a sequential integer value. The RMSE will thus compute how far off a misclassified document is from its correct class. So, for example, a classifier that misidentifies a children's story as a recipe (the next complexity level) will achieve a lower RMSE than one that classifies it as a legal document (the most complex level). \autoref{tab:modelRA} show the RMSE and accuracy score for our baseline and four best models configuration.

Unsurprisingly, the baseline achieves the lowest accuracy and second-worst RMSE. Moreover, as indicated previously and illustrated in \autoref{fig:boxplot_km}, the Kandel and Moles metric gives progressively lower scores to increasing categories of text complexity, and the scores of consecutive categories overlap a lot. As a result, the baseline has trouble pinpointing an exact complexity level for a text.

The DT classifier achieves a better accuracy but worse RMSE than the baseline, and performs worse on both points than the other ML algorithms. DT combines the information of several metrics while the baseline is only one metric, so its higher accuracy compared to the baseline is not surprising. However, its main weaknesses are that it focuses on a subset of the features available, and it applies the same features to sort between all complexity levels at the highest levels of the tree. As our analysis in \autoref{sec:complexity} will show, different features are indicative of different complexity levels, and almost every feature can come into play. This makes DT unsuited for this task, and explains its lower performance compared to the other algorithms.

The LR, NB and RF classifiers all achieve very strong accuracy and RMSE scores in our experiments. The LR classifier is built as a set of eight individual one-vs-rest regression equations, each one computing the score of the document belonging at that complexity level. The NB classifier computes, for each complexity level, the product of conditional probabilities of that complexity given the observed features of the text. The equation yielding the highest score dictates the predicted complexity level in both cases. The RF classifier is composed of an ensemble of 1,244 independent DTs.
Most importantly, both LR and NB consider all features available, and both can weight these features differently for each complexity level independently of the others. Each DT in the RF uses a different random subset of features, but the ensemble will likewise make use of all features. 
This gives these algorithms the flexibility needed to outperform DT and the baseline. The NB classifier also benefits from smoothing the probabilities of unseen events, which is an advantage given the small size of our dataset and allows it to beat out the LR in the RMSE score. The RF, on the other hand, benefits from merging the decisions of an ensemble of individual trees, each trained to test their subset of features optimally. This flexibility allows it to beat out the other ML algorithms and achieve the best accuracy and RMSE.

It is interesting to look more closely at the use of the features made by ML classifiers. 
With the LR, normalized metrics that have large positive or negative coefficient values are the ones that have the most influence on the score of a level. In contrast, those with values near zero have little to no influence. Moreover, metrics whose coefficients vary significantly from one equation to the other will have a greater ability to differentiate one complexity level from another. In comparison, metrics whose coefficient values remain constant from one equation to another have no discerning power. 
Using this standard, we can discern many significant metrics that distinguish between two or more levels of complexity. The TTR, for instance, has a very high positive value for Wikipedia articles and a very high negative value for legal documents, owing to the former's linguistic variety and the latter's repetition of formally-defined terms. Conjuncts (the use of conjunctions) is also a strong positive indicator of the dictations and a strong negative indicator of recipes, showing the former class' reliance on syntactic complexity to challenge learners and the latter's use of clear and simple statements. 
There are also metrics that distinguish groups of levels. For instance, contractions have a strong positive coefficient for children's stories and dictations, and a strong negative coefficient for Wikipedia articles, novels, and legal documents. 
Some metrics are strong indicators for only one complexity level and null for others. For example, this is the case for the use of past-tense verbs (strong positive coefficient for insurance contracts) or the use of adverbs (strong negative coefficient for legal documents). 
Finally, a few metrics had near-zero coefficients in all LR equations. This was notably the case for the use of second-person pronouns, pied-piping, suasive verbs, time adverbs, or the use of seem/appear.

A similar observation holds true for the NB. 
Since our metric values are normalized, a significant indicator for a complexity level metric will have a high conditional probability. Moreover, if a metric is a significant indicator of one complexity level but not of another, the conditional probability of the first level, given that metric will be high and that of the second level, will be low.
By that standard, the most significant metrics according to NB include the use of contractions, syntax negation and 3rd person pronouns.
The first metric is a high probability for children's stories, recipes and dictations but a low probability for novels and legal documents.
The second is a high probability for insurance and legal documents but a low probability for Wikipedia articles.
The third metric is a high probability for stories, news articles and Wikipedia articles but a low probability for recipes and insurance contracts. 
On the other hand, insignificant metrics that have uniformly low probability values for all or almost all complexity levels include pied-piping, suasive verbs, time adverbs and the use of seem/appear. All of these also had coefficients near zero in all LR equations.
Equally interesting are metrics that have uniformly high probability values for most complexity levels, the most striking of which are the FK ease and KM score. These reading ease metrics are correlated with text complexity but overlap in values for neighbouring complexity levels, as was shown in \autoref{sec:resultsmetric}. This seems to keep the LR from using them, but does not hinder the NB.

Unlike the LR and NB, the DT and RF create one classifier for all complexity levels. This means that we can weight the importance of each metric to the classification, but we cannot determine which metric is significant for each complexity level. We find the DT uses 15 metrics, while the RF unsurprisingly uses all of them. More interestingly, when sorted by importance, the order of metrics roughly agree, with more or less important metrics picked by the DT weighted as more or less important by the RF as well. The most important features agreed upon by both classifiers are the TTR and MTLD scores, the $\frac{\text{CP}}{\text{C}}$ and $\frac{\text{DC}}{\text{C}}$ ratios, the use of impersonal pronouns and nominalizations, and the KM score. Many of these metrics are also rated as strong indicators of various complexity levels by the LR and NB classifiers as well.

\subsection{Assessing French Text Complexity}
\label{sec:complexity}

Overall, it seems the ML classifiers isolated, for each class of documents, a set of metrics that describe what characteristics documents of that class typically have or do not have. Since we selected the eight classes to represent eight different levels of text complexity, these characteristics should also match those that make documents easier or more difficult to read. Following our discussion of \autoref{sec:resultslearning}, we decided to focus on the LR classifier. While it does not perform as well as the NB or RF in \autoref{tab:modelRA}, its performances are still good, and more importantly it is directly interpretable from the coefficients of the features in each equation.
\autoref{tab:complexfeatures} presents the list of metrics that receive strong positive or negative coefficients, which we define as an absolute value above 0.1, for each complexity level. 
Looking at those results makes it possible to see features of text complexity emerge.
\begin{itemize}[itemsep=2pt]
  \item Legal documents, the most complex form of text in our study, represent a form of syntactic complexity that uses long and complex sentence structures with T-Units, dependent clauses, coordinate phrases, and commas. The vocabulary is more formal, with low diversity and little use of contractions and adverbs.
  \item Insurance contracts, the other most complex form of text, represent instead a form of grammatical complexity, making use of modal verbs, long words, and stranded prepositions.
  \item Dictations show a form of vocabulary complexity, including nouns, adjectives, contractions, and rare words, but simpler sentences with \textit{to be} as the main verb.
  \item Novels, one of our intermediate levels of text complexity, represent a more sophisticated prose, with prepositions, causative verbs, downtoners and emphatics to modify other words, and a lack of coordinate phrases.
  \item Wikipedia articles represent a more technical form of prose, with a high vocabulary variety, longer words, and few contractions, but simple sentences using verbs like \textit{to be} and \textit{to do}.
  \item News articles are the simplest of our intermediate complexity levels, characterized by vocabulary diversity and a heavy use of pronouns.
  \item Recipes, the first of our two easiest levels of difficulty, are characterized by short sentences that avoid any sophistication such as \guillemet{wh-} subordinate clauses, conjunctions, questions, and \guillemet{that} deletion.
  \item Children stories are the simplest level of text complexity in our study, and are characterized by the use of simplifying acts such as that-deletions and contractions, and the use of shorter words and short sentences.
\end{itemize}

Another interesting observation is that most of our metrics are useful in modelling text complexity. Indeed, we had included several related metrics in our study, which measure the same characteristics but with slight variations. We expected to see some of them retained by the ML algorithms and others ignored, but this turned out not to be the case. For example, we included four variations of the TTR metric in \autoref{tab:ld_metrics}, and all four were used to identify some complexity levels. It indicates that the characteristics that make a text easier or harder to read can be very pointed, and thus the differences between these variations of metrics capture useful information.

\begin{table}
    \centering
    \captionsetup{width=\linewidth}
    \begin{tabular}{p{0.125\linewidth}p{0.475\linewidth}p{0.4\linewidth}}
    \toprule
        \textbf{Complexity} & \textbf{Positive Coefficient} &  \textbf{Negative Coefficient}\\
        \midrule
        Legal document & BINGUI, $\frac{\text{C}}{\text{S}}$, $\frac{\text{CP}}{\text{C}}$, $\frac{\text{CP}}{\text{TU}}$, $\frac{\text{DC}}{\text{C}}$, MLS, nominalizations, possibModals, $\frac{\text{TU}}{\text{S}}$ & ADV, TTR, contractions, KM score, MSTTR, $PS_{30}$ \\ \\
        Insurance document & $\frac{\text{C}}{\text{TU}}$, conditional, indefProns, necessModals, NLM, pastVerbs, predicModals, strandedPrep, syntNegn, wordLength  & (none) \\ \\
        Dictation & Nouns, attrAdj, beAsMain, conjuncts, contractions, NLM & $\frac{\text{C}}{\text{TU}}$, indefProns, unigram \\ \\
        Novels & 1persProns, analNegn, causative, $\frac{\text{CTU}}{\text{TU}}$, downtoners, generalEmphatics, MLT, preposn, publicVerbs  & 3persProns, contractions, $\frac{\text{CP}}{\text{C}}$, $\frac{\text{CP}}{\text{TU}}$ \\ \\
        Wikipedia article & TTR, beAsMain, $\frac{\text{CTU}}{\text{CU}}$, doAsProVerb, MSTTR, wordLength & analNegn, causative, conditional, contractions, preposn, sncRelatives \\ \\
        News article & 1persProns, 3persProns, analNegn, demonstrProns, impersProns, MATTR, MTLD & NLM \\ \\
        Recipe & (none) & WHclauses, attrAdj, beAsMain, conjuncts, MLT, nominalizations, pastVerbs, presVerbs, thatDeletion, whQuestions\\ \\
        Story & 3persProns, amplifiers, contractions, impersProns, placeAdverbials, privateVerbs, thatDeletion & MLT, PA, wordLength \\
        \bottomrule
    \end{tabular}
    \caption{Metrics with strong positive and negative coefficients (LR classifier) for each text complexity level}
    \label{tab:complexfeatures}
\end{table}

\subsection{Blind Tests}
\label{sec:validation}
We decided to run a final test of our methodology. While our classifiers work well to recognize the complexity levels of documents belonging to the categories seen during training, we now want to see if they can also handle documents of related complexity levels but of different categories or genres. To this end, we picked six new categories of documents and collected 5 sample texts for each. They are: 
\begin{itemize}[itemsep=2pt]
\item movies subtitles, documents consisting of short simple sentences akin to children stories and recipes and which should thus rank at levels 0 or 1;
\item tweets, also simple texts but from a very different source than the previous;
\item introductory-level college course materials, informative documents aimed at non-specialized learners. They should be a bit more complex than Wikipedia or news articles, which are informative documents aimed at the general public, at level 4;
\item philosophical essays, the most different type of documents from our original corpus, but which we expect to be around complexity levels 4 or 5, meaning they would be more complex than general-public-level technical documents but less complex than specialized legal documents;
\item websites terms of services, which are equivalent to insurance contracts at level 6;
\item laws from French-speaking European countries, equivalent to level 7.
\end{itemize}
We then classified each document independently into a complexity level using the four classifiers we trained in \autoref{sec:resultslearning}, and computed the average. The resulting detected complexity levels are given in \autoref{tab:my_label}. 

\begin{table}
    \centering
    \begin{tabular}{lccccc}
    \toprule
        Category & Expected & DT & LR & NB & RF  \\
        \midrule
        Movies subtitles & 0-1 & 0.0 & 3.6 & 0.0 & 2.0 \\
        Tweets & 0-1 & 2.0 & 3.2 & 1.2 & 3.6 \\
        College materials & 4 & 4.8 & 4.8 & 3.8 & 4.5 \\
        Philosophical essays & 4-5 & 4.0 & 5.4 & 1.8 & 3.6 \\
        Terms of Service & 6 & 5.2 & 6.0 & 6.0 & 5.4 \\
        Foreign laws & 7 & 5.0 & 6.0 & 5.8 & 4.8 \\
        \midrule
        Correlation to expected levels & & 0.90 & \textbf{0.98} & 0.90 & 0.82 \\
        \bottomrule
    \end{tabular}
    \caption{Predicted complexity levels of the blind test document set} 
    \label{tab:my_label}
    \vspace{-2em}
\end{table}
The results show some interesting patterns. To begin, almost all algorithms overestimate the complexity levels of subtitles and tweets. It may be because the two least complex classes we trained with, recipes and news articles, were marked by low vocabulary diversity, which is not the case for either of these test classes. In fact, tweets, with their frequent use of spelling mistakes, hashtags, usernames and pop-culture terms, have a very high vocabulary diversity, which misleads most classifiers into ranking them near the middle of our complexity levels. 
Likewise, the massive use of pronouns in subtitles, which are transcripts of spoken discussions, is a typical trait of the news article class and leads to overestimating their complexity by two classifiers.
On the other hand, the complexity of the foreign law class is very under-estimated by all four classifiers. This is because French-Canadian laws, which form our level-7 complexity (highest level), are very different from European laws, which appear more akin to Canadian insurance contracts. 
For example, Belgian laws are structured as short one-sentence articles, whereas Canadian ones use complex compound articles with lists of cases.
On the positive side, however, the classifiers were good at pinpointing the complexity levels of college course materials, philosophical essays (except for NB), and terms of service documents. 

More importantly, it can be seen that all four classifiers respect the order of rankings of complexity levels. They all find the Tweets and subtitles to be the simplest levels, the terms of service and foreign laws to be the most complex levels, and the college materials and philosophical essays of intermediate complexity between them. Moreover, the correlation coefficient between our expected complexity levels and those produced by each of the classifiers ranges between 0.82 and 0.98. They thus classify text complexity in a manner coherent with our intuition, with the LR ranking complexity in an order closest to ours. This observation confirms that our methodology picks out document complexity and does not simply classify documents by type or topic.
\vspace{-0.5em}

\section{Conclusion}
\label{sec:conclusion}
In this paper, we studied the concept of written text complexity in French. To this end, we built a corpus of 118 French documents covering 8 levels of complexity ranging from children's stories to insurance contracts, and selected some 60 different popular metrics of text complexity. Our first realization is that none of these metrics can, by itself, correctly model the range and diversity of text complexity. Consequently, we applied a variety of ML algorithms, and found that logistic regressions, Naive Bayes, and random forests all can discover good combinations of metrics to predict text complexity, with RF edging out the other two in performance while the LR is most explainable. Taking advantage of that explainability, we were then able to infer the set of features that define each complexity level. This in turn made it possible to apply our algorithms successfully to new types of documents and correctly measure and rank their complexity, with LR giving the most human-like rankings. Our corpus, code and results are all available on our GitHub repository.

Our proposed method is, to the best of our knowledge, the only general-purpose algorithm to accurately measure a wide range of text complexity in French. Moreover, the explainability of the method gives insight into the nature of the French language as well as confidence in the predicted complexity level of a document. Finally, while we did not experiment in that direction, we believe our methodology can be applied to different languages as well, by creating a corpus in the new language and possibly adding new metrics to model unique aspects of that language's grammar. 

\section*{Acknowledgements}
This research was made possible thanks to the support of \textit{La Capitale Assurances et Services Financiers} and NSERC research grant RDCPJ 537198-18.

\printbibliography[heading=subbibintoc]

\end{document}